\documentclass{article}

    \usepackage[final]{neurips_2024}

\usepackage[utf8]{inputenc} 
\usepackage[T1]{fontenc}    
\usepackage{hyperref}       
\usepackage{url}            
\usepackage{booktabs}       
\usepackage{amsfonts}       
\usepackage{nicefrac}       
\usepackage{microtype}      
\usepackage{xcolor}         
\usepackage{amsmath}
\usepackage{amssymb}
\usepackage{graphicx}
\usepackage{subcaption}
\usepackage{float}
\usepackage{amsmath}
\usepackage{bm}
\usepackage{natbib}
\usepackage{booktabs}
\usepackage{multicol, multirow}

\usepackage{listings}
\usepackage{xcolor} 

\title{The N-Grammys: Accelerating Autoregressive Inference with Learning-Free Batched Speculation
}

\author{%
  Lawrence Stewart\thanks{Work done during an internship at AWS AI Labs.} \\ ENS Paris \\
  \And
  Matthew Trager \\
  AWS AI Labs \\
  \And
 Sujan Kumar Gonugondla\thanks{Work done while at AWS NGDE Science.} \\
  Meta AI \\
  \And
  Stefano Soatto \\
  AWS AI Labs \\
}

\newcommand{\eg}{{\em e.g. }}
\newcommand{\ie}{{\em i.e. }}

\begin{document}

\maketitle

\begin{abstract}
Speculative decoding aims to speed up autoregressive generation of a language model by verifying in parallel the tokens generated by a smaller draft model.
In this work, we explore the effectiveness of learning-free, negligible-cost draft strategies, namely $N$-grams obtained from the model weights and the context. While the predicted next token of the base model is rarely the top prediction of these simple strategies, we observe that it is often within their top-$k$ predictions for small $k$. Based on this, we show that combinations of simple strategies can achieve significant inference speedups over different tasks. The overall performance is comparable to more complex methods, yet does not require expensive preprocessing or modification of the base model, and allows for seamless `plug-and-play' integration into pipelines.
\end{abstract}

\section{Introduction}

Large Language Models (LLMs) have had a significant impact across various scientific and industrial domains. However, their autoregressive decoding process, which generates one new token per model call, is computationally expensive. This issue is particularly challenging for larger models, which typically exhibit superior performance compared to smaller ones~\citep{gpt3_2020, PALM2_2023, gpt4_2023}.

To improve computational efficiency and inference latency, many works have proposed methods for reducing the cost of a single model call. Some examples include quantization methods~\cite{Yao2023ACS}, early-exiting strategies~\cite{Xin2020DeeBERTDE}, flash-attention \citep{dao2022flashattention, dao2023flashattention}, and multi-token prediction \citep{gloeckle2024multitoken}. 

Another line of work has considered variants of autoregressive decoding aimed at better leveraging the parallel processing capabilities of GPU/TPU hardware accelerators. In particular, \emph{speculative decoding} methods \citep{stern2018blockwiseparalleldecodingdeep,specdec1, specdec2, specdec3}, sometimes also called \emph{guess-and-verify} methods, use a smaller ``draft model'' to generate proposals for multiple future tokens. They then ``validate'' all of these tokens in parallel with a single call to the original model and ensuring that the original model would have predicted the same tokens. This approach is similar to speculative execution \citep{specexec}, in which a processor executes instructions in parallel to verifying if they are needed, trading resources for concurrency, as in just-in-time XLA compilation in JAX, Pytorch, and Tensorflow.

The effectiveness of speculative decoding is determined by i) the acceptance rate of the draft speculations, ii) the discrepancy in call-time between the original model and the draft model, and iii) the extra cost required for parallel verification by the main model (although this is often assumed to be negligible when using hardware accelerators).

Choosing a draft model that is compatible with the base model and within the available compute budget can be challenging. To address this issue, several approaches have been developed that involve augmenting the base model and performing supervised fine-tuning (SFT), with the goal of ensuring that the draft model and the base model utilize the same feature representations \citep{cai2024medusa, li2024eagle, bhendawade2024streaming}. Although these strategies often achieve high acceptance rates, they come with the drawback of necessitating SFT for each individual model, which can be resource-intensive.

To overcome this, researchers have also considered \emph{negligible cost draft models}, remarking that if the cost of the draft model is close to zero then even a low acceptance rate can yield wall-time speedups. This was first proposed explicitly in \citet{specdec1}, where the authors experimented with unigram and bigram draft models, both trained on external data, as well as implicitly by \citet{jacobi}, who used a cost-free draft model, corresponding to the base model's greedy predictions at a previous time step.

In this paper, we aim to explore the full potential of negligible-cost draft methods for accelerating autoregressive decoding. In particular, we argue that even simple strategies --- based on N-grams derived from the model and the context --- can be very effective when combined in a batch to explore the space of possible future trajectories in parallel. Our proposed methods have the following desirable features: \textbf{(P1)} they do not require training a draft model or finetuning the base model, \textbf{(P2)}  they make use of no external data or external draft model, and most importantly \textbf{(P3)}  they can easily be integrated with any existing pipeline as an \emph{out-of-the-box} approach, and moreover they can be combined with other acceleration techniques like the ones mentioned above.

We emphasize, the goal of this paper is not to achieve state-of-the-art inference speed-ups but rather explore the strengths and weaknesses of simple methods that satisfy the desirable properties \textbf{(P1)}, \textbf{(P2)}, \textbf{(P3)}. Our experiments across different datasets (MTBench~\cite{zhengJudgingLLMasaJudgeMTBench2023}, HumanEval~\cite{chenEvaluatingLargeLanguage2021}, GSM8K~\cite{cobbeTrainingVerifiersSolve2021}) and models (Mistral7B~\cite{jiangMistral7B2023}, Phi-3~\cite{abdin2024phi}, Vicuna13B~\cite{zhengJudgingLLMasaJudgeMTBench2023})  show that such \emph{out-of-the-box} strategies are surprisingly effective.

\paragraph{Main contributions.}
\begin{enumerate}
    \item We first revisit the critical assumption made by guess-and-verify methods, namely memory-bound hardware parallelism, highlighting situations where this assumption may fail. 
    \item We present a class of learning-free strategies for generating batches of speculative drafts with negligible computational cost. These strategies are model-independent and can be implemented with minimal wrapper-code, enabling easy integration into existing systems.
    \item We conduct an in-depth analysis and evaluation of our methods, demonstrating that combining simple strategies can lead to substantial speedups across a diverse set of tasks.
\end{enumerate}

\section{Further related work}

\paragraph{External draft model.} The concept of guess-and-verify using an \emph{external draft model} was explicitly proposed in several concurrent works  \citep{specdec1, specdec2, specdec3}, with \citet{specdec1} notably exploring negligible cost models and suggesting that fitting N-grams to the context could potentially be a promising line of future work. Recently, follow-up works have investigated using a collection of varied size draft models \citep{chen2023cascade}, tree-based guess-and-verify methods \citep{miao2023specinfer}, retrieving speculations from external data sources \citep{he2023rest}, and using a collection online-buffers for aligning the draft and base model via training \citep{liu2023online}. Contrary to these works, we aim to explore strategies that do not require an external draft model.

\paragraph{Learning by adapting the base model.} In order to align the predictions of the draft model with those of the base model, several works have proposed grafting a draft model on top of the base model, so that both share the same features. \citet{cai2024medusa} propose adding $K$ heads to a model in order to predict $K$ tokens into the future, together with a tree-based attention mechanism. The authors explore i) fine-tuning only the heads with the base-model frozen ii) fine-tuning the base LLM with the heads. Building on this idea, \citet{li2024eagle} proposes training an auto-regressive decoder from the penultimate layer, showing that this approach can obtain very high acceptance rates.

\paragraph{Learning-free methods.} \citet{jacobi} proposed initializing a random speculation, and at subsequent decoding steps using the model predictions from the previous step as speculations, in order to improve upon greedy decoding. This process resembles the Jacobi and Gauss-Seidel iterative methods (and is hence called Jacobi Decoding), and is implementable just a few lines of code. Look-ahead decoding \citet{fu2024lookahead} further improves upon the acceptance rate of Jacobi decoding, by using a custom-attention mask to generate an N-gram speculation cache as well as verifying matching speculations in parallel. 

\subsubsection*{Glossary}
\begin{table}[h!]
\small
    \centering
    \vspace{-1.5em}
    \begin{tabular}{ll}
        \toprule
        \textbf{Symbol} & \textbf{Usage} \\
        \midrule
        $\mathcal{X}$ & the set of tokens that constitute the vocabulary of an LLM.\\
        $k$ & number of batched speculations, taken from the top-$k$ of probability over tokens.\\
        $w$ & number of tokens speculated into the future. \\
        $q$ & number of query tokens to match with context when looking for an 
$N$-gram. \\
        $\ell$ & length of context at a decoding step, assumed to be key-valued cached (KV-cached). \\
        \bottomrule
    \end{tabular}
    \label{tab:glossary}
        \vspace{-1.5em}
\end{table}
\paragraph{Limitations.} For simplicity and ease of integration, our method incurs extra computation cost through batching (which could be addressed in follow-up works by incorporating methods such as bifurcated attention~\cite{athiwaratkunBifurcatedAttentionSingleContext2024}). Further exploration is needed for non-greedy sampling methods such as those discussed in \citep{specdec1}, which are commonly deployed. Finally, we have limited our experiments to decoder-only transformer models, and it remains to be tested with other architectures such as state-spaced models \citep{mamba}.

\section{Assumptions on parallelism for verification}
We briefly revisit and clarify the key assumption in the \emph{guess-and-verify} literature \citep{specdec1, specdec2, specdec3, jacobi} that parallel verification of speculated tokens by the base model is memory-bound when using hardware accelerators like GPUs/TPUs.

Hardware accelerators like GPUs and TPUs divide matrix multiplications (matmuls) into tiles, each assigned to independent computation threads. These operations can be parallelized if their operations-to-bytes ratio (OTB) is below the hardware's threshold, allowing all threads to be assigned to multiprocessors and executed concurrently. If the OTB ratio is above the threshold, the operation becomes compute-bound (or math-bound), as the number of tiles exceeds the number of multiprocessors, requiring the total number of tiles to be quantized to fit the available hardware resources.

For a fixed model and hardware, let $\ell$ denote the length of the given context at any decoding step, with all context tokens (except for the final token), assumed to have KV-cache stored in memory. Let $(k, w + 1)$ denote the dimensions of the input batch, where $k\geq 1$ denotes the batch size and $w \geq 0$ denotes the number of tokens speculated into the future. With this terminology, the \emph{guess-and-verify} assumption above can be rephrased as the following: `\textit{for a fixed model using KV-caching and a fixed hardware accelerator, and for a given context length $\ell \geq 1$, the time required to perform a model call on an input block of size $(k, w+1)$ is approximately the same as the time for a model call on an input block of size $(k, 1)$.}

\begin{figure}[t]
  \centering
  \begin{subfigure}[b]{0.323\textwidth}
    \includegraphics[width=\linewidth]{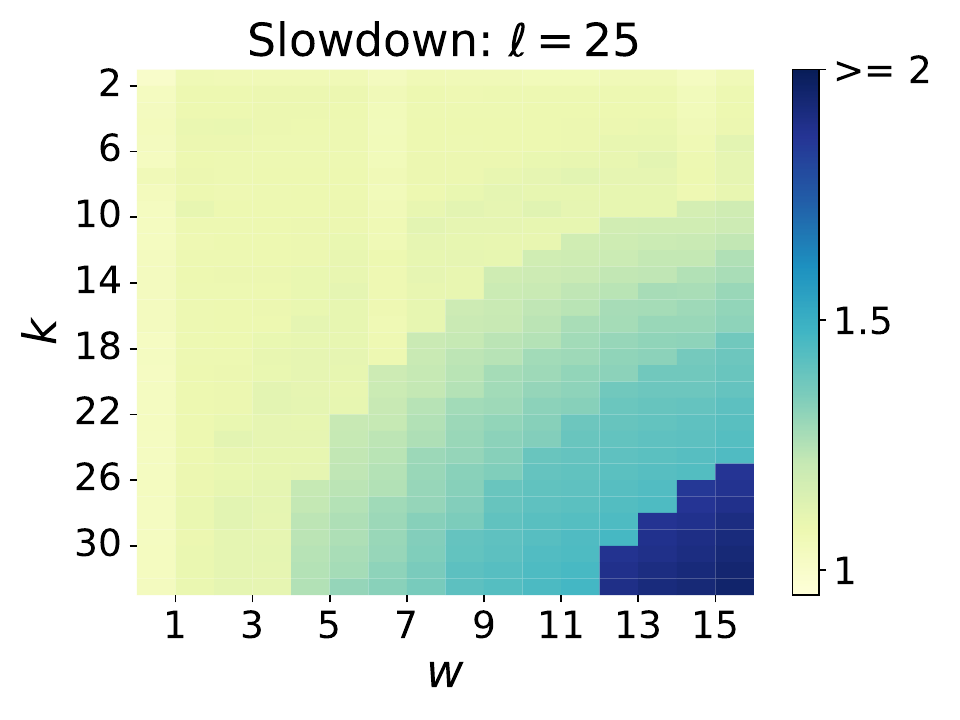}
    \label{fig:sub1}
  \end{subfigure}
  \begin{subfigure}[b]{0.323\textwidth}
    \includegraphics[width=\linewidth]{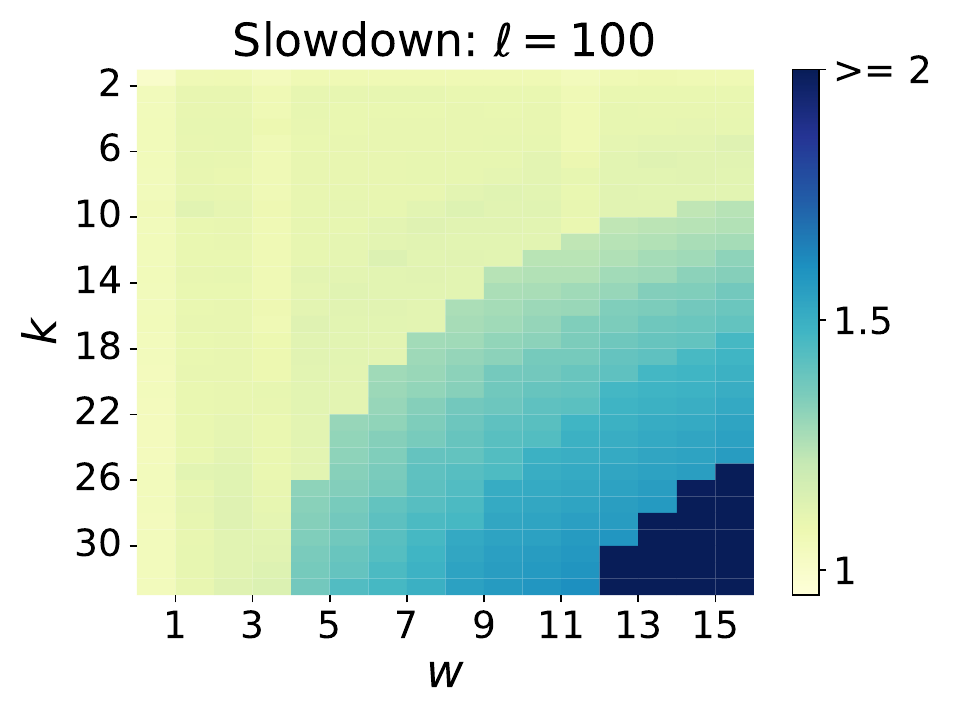}
    \label{fig:sub2}
  \end{subfigure}
  \begin{subfigure}[b]{0.323\textwidth}
    \includegraphics[width=\linewidth]{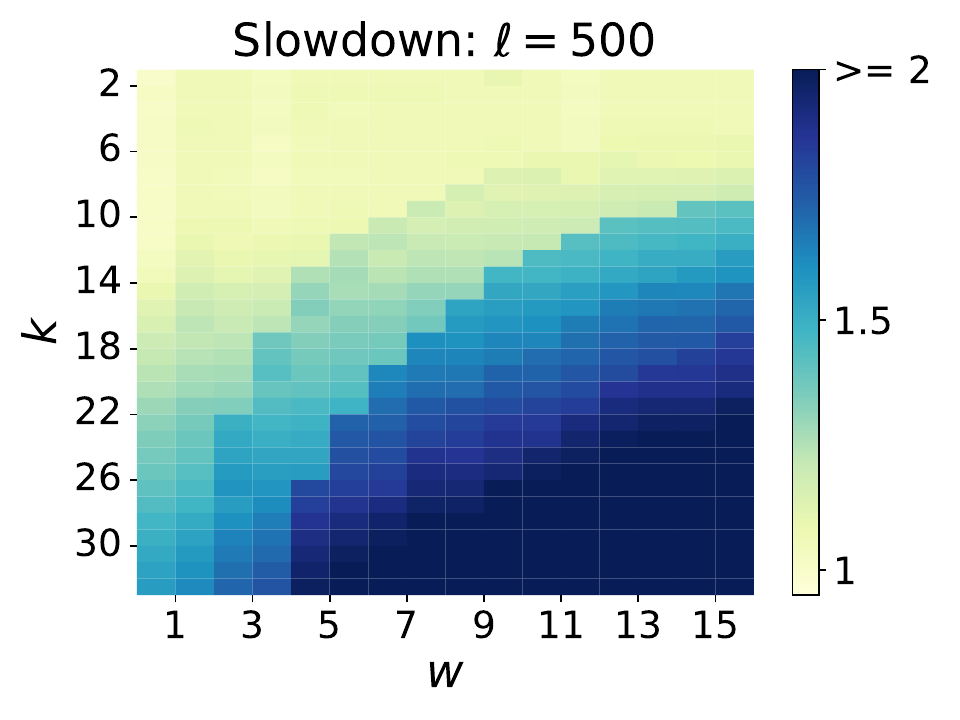}
    \label{fig:m7b-slowdown}
  \end{subfigure}
  \caption{\textbf{Memory-bound to compute-bound transition:} The heatmaps depict the slowdown of a model call, for varied batch size $k\in \{1, \ldots, 32\}$ and speculation length $w \in\{0, \ldots, 15\}$. The slow-downs are relative to that of standard greedy decoding with no speculation \ie $(k, w)= (1, 0)$. The leftmost plot corresponds to a context-length of $\ell=25$, the middle to $\ell=100$ and the rightmost plot to $\ell=500$. The model used was Mistral 7B at standard bfloat-16 precision, with a single NVIDIA A100 GPU with 40GB of memory. Each square in the heat-maps corresponds to the average slow-down over five model calls.}
  \label{fig:slowdown}
\end{figure}

When does this assumption hold? For each element in a batch of size $k$, the attention mechanism requires multiplying $(w+1)$ queries by $(\ell  + w)$ keys, resulting in $\mathcal{O}\left(kw(w+\ell)\right)$ complexity. So for a fixed model, accelerator and $(\ell, k, w)$ values, the assumption holds if and only if all matmuls in the forward pass of the model have an OTB ratio less than the accelerator's threshold. In practice, this does not always hold. Figure ~\ref{fig:slowdown} depicts the phase-transition from memory-bound to compute bound, with varied  $(\ell, k, w)$, for Mistral 7B on a NVIDIA A100 40GB GPU. One does not see a smooth scaling of $\mathcal{O}\left(kw(w+\ell)\right)$ in the phase transition, due to the quantization to multiprocessors, resulting in jumps known as \emph{wave quantization}.  

In the special case of neglible cost draft models, for which the time to generate speculations is assumed to be near zero, there will be a clear trade-off between the speed-up gained from accepting extra tokens (by increasing $k$ and $w$) vs the potential slow-down that could be faced when entering a compute-bound setting, where the guess and verify assumption is broken.

\section{Learning-free drafting}

The use of N-grams to model language dates back at least to \citet{markov2006example}, who published a paper in which he used conditional probabilities of constants and vowels (computed by hand) to compare the poem \textit{Eugene Onyegin} by Pushkin to other texts, showing how unigram and bigram probabilities could mathematically capture an author's style. Many sequences of tokens in natural language / computer code exhibit low entropy, making even simple N-grams effective at predicting them, as demonstrated in \citet{shannon1951prediction}. In this section, we explore ways to extract N-grams directly from a large language model and a context, and use these for speculation. These methods are learning-free, as they require no training \textbf{(P1)}, nor external data \textbf{(P2)}, in contrast to the N-gram models used in \citet{specdec1}, which are obtained from external data sources.  All the methods discussed in this section can be implemented with minimal wrapper code (detailed in Appendix \ref{app:Ngrams}), allowing for them to be added to existing pipelines with minimal friction \textbf{(P3)}.

\subsection{Model-derived N-grams}

Let $M$ denote a language model with vocabulary $\mathcal{X}$. Let $V\in \mathbb{R}^{\mathcal{X} \times d}$ and $U\in \mathbb{R}^{d \times \mathcal{X}}$ denote the model's input and output embedding layers, with respective row / column word embeddings $\{v_i\}_{i\in | \mathcal{X}|}$ and $\{u_i\}_{i\in | \mathcal{X}|}$. For a given context $c$ (i.e. a sequence of tokens from $\mathcal{X}$), we denote the next token distribution according to $M$ as $p_M(\cdot | c)$, where $\sum_{x \in \mathcal{X}} p_M(x | c) = 1$.

\paragraph{Unigram.} 

Consider the function $d(x) = \| u_x - \bar{u} \|_{V}$, where $\bar u\in \mathbb{R}^d$ is the mean token output embedding, and $\|\cdot \|_{V}$ is the distance induced by the covariance matrix of the input embeddings $V$, that is, the inner product $\langle u_1, u_2\rangle_{V} = u_1^T V^T V u_2 = \sum_{x \in \mathcal X} (u_1^T v_x) (u_2^T v_x)$. This product is more natural than the standard one in $\mathbb R^d$ as two tokens will be close when they lead to similar distributions for the following token for the model, as captured by the vectors $(u_i^T v_x \colon x \in \mathcal X) \in \mathbb R^{|\mathcal X|}$. We can hence define a unigram distribution over tokens using the input and output embeddings as $p(x)\propto e^{-d(x)}$.

\paragraph{Bigram.} We can easily obtain a bigram model from a language model $M$ by calculating $p_M(\cdot \, | \, x)$ for all tokens $x \in \mathcal X$. 
For typical models, this can be a calculated once for every $x\in \mathcal{X}$ and stored for quick use later. For example, generating such a bigram model takes $\leq 1$ minute for Mistral 7B on a single A100 GPU, and is a one-off cost. While this simple bi-gram lacks context for tokens prior to $x$, it can still be effective, particularly in cases where the preceding context is not essential for making accurate predictions.

\paragraph{Batched drafts.} When performing autoregressive decoding, the greedy next token prediction (NTP) of the base model will seldom match with that predicted by the above unigram and bigram models (derived from the base model). However, we remark that the base model NTP appears often amongst the top-$k$ predictions of the N-grams, even for small $k$. Consequentially, we propose obtaining speculations from the top-$k$ of a N-gram model, i.e., $s : \mathcal{X}^q \rightarrow \mathcal{X}^{k \times 1}$, where $q \geq 0$ denotes the number of last context tokens to use to produce the speculation (i.e. $q=0$ for the unigram and $q=1$ for the bigram), and the speculation $s$ returns the top-$k$ \textit{next-word} token predictions according to the N-gram.

Speculative decoding using our model derived unigram / bigram can be easily implemented in the following manner: i) repeat the context\footnote{If the context has a key-value cache, one can utilize a method to obtain a batched tensor view without consuming additional memory, e.g., \textit{torch.extend}.} to form a batch of $k$ identical rows; ii) append a column corresponding to the top-$k$ speculations to the end of the batch; iii) call the model on the batch to verify all speculations (rows) in a single forward pass.

While adding redundant computation (regarding repeat flops for the context), this implementation is \emph{extremely simple} to integrate into existing code, and in addition, is fully-compatible with popular inference methods, \eg, flash-attention \citep{dao2022flashattention, dao2023flashattention} and paged-attention / vLLM \citep{vllm}, which is generally not the case for methods that require custom attention masking such as tree / lookahead attention masking \citep{fu2024lookahead,cai2024medusa}.

\begin{figure}[!t]
  \centering

  \begin{subfigure}[b]{0.43\textwidth}
    \includegraphics[width=\linewidth]{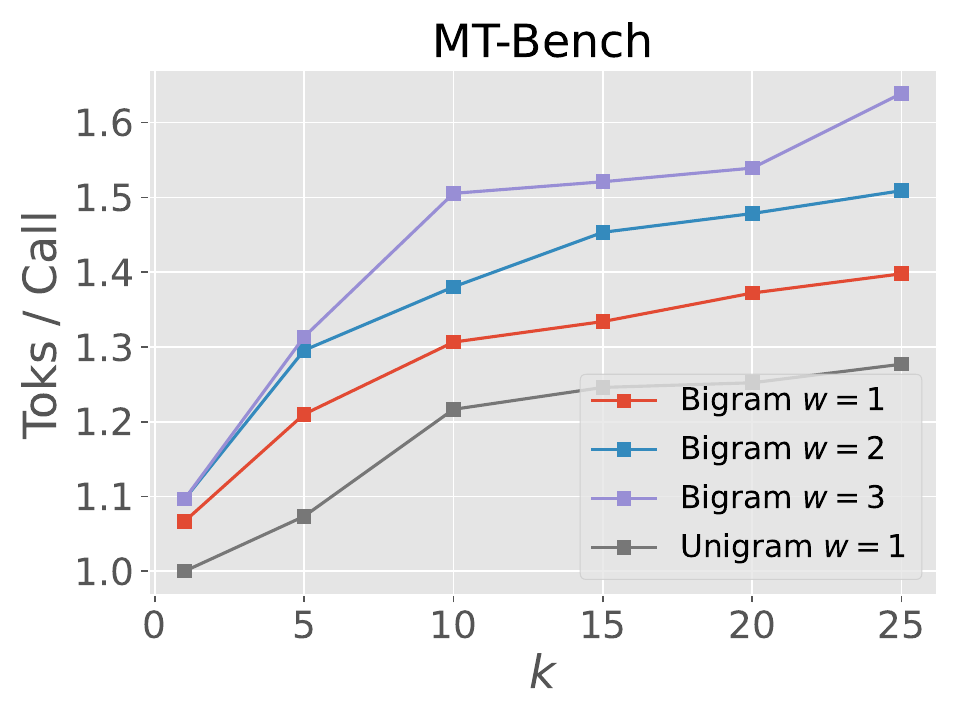}
  \end{subfigure}
  \hfill
  \begin{subfigure}[b]{0.43\textwidth}
    \includegraphics[width=\linewidth]{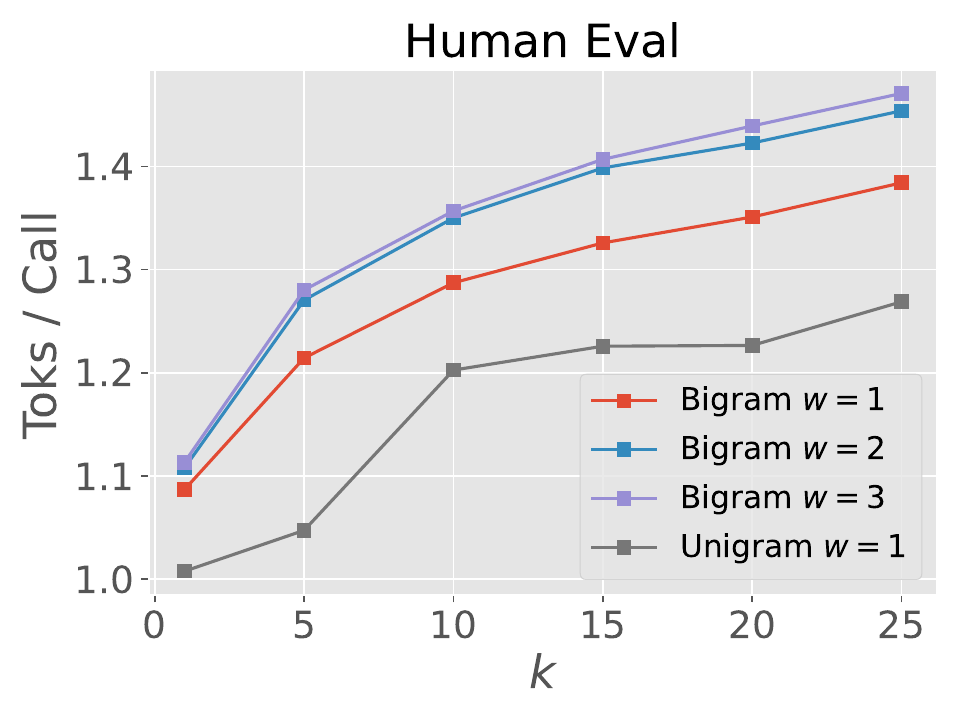}
  \end{subfigure}
  \caption{Tokens per call as a function of $k$, the top-$k$ speculations of the model derived unigram / bigram. In addition, the plot depicts the extended bigram (described below) plotted for $w=2$ and $w=3$, showing gains comparing $w=1$ to $w=2$, but diminishing gains going to $w=3$. The results were obtained on the first 50 examples of MT-bench and Human Eval, using a 7B model (Mistral Instruct) 
\citep{jiangMistral7B2023}.}
  \label{fig:tokscale}
\end{figure}

\paragraph{Extensions.} The model-derived bigram and unigram allow for speculating $w = 1$ token into the future. In addition, by repeatedly applying either the model bigram (or, alternatively greedily, decoding with the model from the bigram), we can easily extend the model-derived N-grams to speculate $w > 1$ tokens into the future $\Tilde{s}: \mathcal{X}^q \rightarrow \mathcal{X}^{k \times w}$. Just like the model bigram, this extension can be generated quickly one time and stored as a $O(1)$ lookup table. 
\paragraph{Top-$k$ speculation.} Figure~\ref{fig:tokscale} provides evidence motivating our approach, depicting how speculating with the top-$k$ tokens increases the number of tokens per call for a 7B model on the first 50 examples of MT-Bench and Human Eval, for both the unigram and bigram models, obtained directly from the Transformer. We remark that with $w=2$, speculating with the top-25 of the model bigram gives a $\approx 50\%$ increase in tokens per call.

\subsection{Context-derived N-grams}\label{subsec:contextngram}

Another natural idea to obtain N-grams for speculations is to look within the context provided to a model; indeed this was suggested by~\citet{specdec1} as a potential avenue for future work. We propose looking for all previous occurrences of the last $q\geq 1$ tokens of the context, and speculating with the $w\geq 1$ tokens that follow a match. To define a discrete probability distribution, we can assign each match a count \ie how many times it occurred in the context, with ties being decided by which match occurred later in the context (hence prioritizing more recent matches). For more details on the context N-gram please refer to the attached code in Appendix \ref{app:contextNgrams}.

\subsection{Mixed strategies}\label{subsec:mixedstrats}

For a chosen batch size $k > 1$, one has the flexibility of using any combination of ``strategies'' \ie model / context derived N-grams, to populate the batch of speculations. This allows for the exploration of diverse combinations of speculation methods. In this work we consider the a straightforward way to mix strategies: first, we populate the $k$ drafts by using as many speculations from the context derived N-gram model as possible, depending on how many matches are obtained (possibly zero); then we use a `extended model bigram' to fill in the remaining speculations. This means that the number of speculations allocated to each strategy (context/model bigram) is variable depending on the context at each step of decoding, which we ablate in Section~\ref{subsec:abl}.

\section{Experiments}\label{sec:exps}

\paragraph{Datasets and models.}

We assess our proposed mixed strategies described in Section \ref{subsec:mixedstrats} using the same experimental setup and datasets as \citet{cai2024medusa, li2024eagle}:  MTBench~\cite{zhengJudgingLLMasaJudgeMTBench2023} (a multi-turn question benchmark with many unique tokens), HumanEval~\cite{chenEvaluatingLargeLanguage2021} (a coding benchmark), GSM8K~\cite{cobbeTrainingVerifiersSolve2021}) (mathematical reasoning problems). We experimented with three different instruction-tuned models of various sizes: Phi-3~\cite{abdin2024phi}, Mistral7B~\cite{jiangMistral7B2023} and Vicuna13B~\cite{zhengJudgingLLMasaJudgeMTBench2023}. All models are freely available from the Hugging-Face transformers library, with reference url links detailed in the Appendix \ref{app:modelurls}. All experiments were run on a single Nvidia A100 GPU with 40GB of memory at bfloat-16 precision.
We report two metrics of interest: \begin{enumerate}
    \item \emph{tokens per call}: measures how many tokens are produced in a single model call on average, i.e., the acceptance rate. This would be the observed speed-up if one had both true parallelism and zero cost speculation.
    \item \emph{wall-time speed-up}: this is the physical observed speed-up on the hardware. To obtain accurate timings, we used the CUDA Runtime API and ran all experiments three times, reporting the mean and standard deviation.
\end{enumerate} 

\paragraph{Mixed strategies.} We consider mixed strategies (as detailed in Section \ref{subsec:contextngram}), defined by values $k\in\{1, 5, 10, 20, 25\}$ and $w\in \{2, 4, \ldots, 14\}$, with query length $q=1$\footnote{We experimented with longer query length $q=2$ and $q=3$, but observed a degradation in both speed-up and tokens per call across all data sets and models.} when deriving speculations from the context. The resulting $(k, w)$ grid totals $35$ different strategies to assess, whose performance will be dictated by trade-offs between i) token per call acceptance and ii) potential compute-bound slowdowns.  

For comparative purposes, we include the results reported by lookahead decoding \citep{fu2024lookahead}, an effective \emph{learning-free} guess-and-verify method, using custom-attention masks to grow an N-gram cache in parallel to verifying speculations. Contrary to look-ahead decoding, our method does not require custom attention masks, due to the naive batching (P3), and is hence fully compatible with methods such as flash-attention \citep{dao2022flashattention, dao2023flashattention}, and is comparatively simpler to integrate / implement.

We also include results reported by REST (Retrieval-Based Speculative Decoding) \citep{he2023rest}, a recent approach which also utilizes negligible draft models. REST requires pre-processed databases to retrieve tree-based speculations. In contrast, our method requires no external data, using only the context and model derived N-grams. We note that it is important to exercise caution when forming exact comparisons between methods and models, since additional factors such as hardware\footnote{Lookahead used a GPU with a higher operations-to-bytes (OTB) ratio than our experiments, whilst the experiments of RAST were conducted on a GPU with a lower OTB ratio.}, tokenizer and instruction formatting will impact the observed speedups for all reported methods.

\subsection{Results}

\begin{table}[!t]
\centering
\begin{tabular}{clcccccc}\toprule
                     &                         & \multicolumn{2}{c}{MT Bench} & \multicolumn{2}{c}{Human eval} & \multicolumn{2}{c}{GSM8K} \\\cmidrule(lr){3-4}\cmidrule(lr){5-6}\cmidrule(lr){7-8}
Model Size           & Strategy                & tok/call       & speedup     & tok/call       & speedup       & tok/call     & speedup    \\\cmidrule(lr){1-8}
\multirow{3}{*}{3b}  
                     & Ours $(10,10)$          & 2.17             & $2.01^{\pm 0.02}$              & 2.28            & $2.11^{\pm 0.01}$              & 2.38            & $2.30^{\pm 0.02}$          \\
                     & Ours $(k^*, w^*)$  & $2.36$             & $2.18^{\pm 0.02}$           & 2.51              & $2.34^{\pm 0.02}$            & 2.41            & $2.51^{\pm 0.01}$ 
                     \\\cmidrule(lr){1-8}
\multirow{3}{*}{7b}  
                     & Lookahead     & --             & 1.65        & --             & 2.25          & --           & 1.89        \\
                     & REST             & --             & 1.69        & --             & 2.12          & --           & --         \\
                     & Ours $(10,10)$          & 2.13           & $1.91^{\pm 0.01}$        & 2.22           & $2.04^{\pm 0.01}$          & 2.16         & $2.03^{\pm 0.03}$          \\
                     & Ours $(k^*, w^*)$  & 2.13           & $1.91^{\pm 0.01}$        & 2.19           & $2.05^{\pm 0.01}$          & 2.16         & $2.03^{\pm 0.03}$         \\\cmidrule(lr){1-8}
\multirow{3}{*}{13b} 
                     & Lookahead     & --             & 1.51        & --             & 2.26          & --           & 1.72         \\
                     & REST           & --             & 1.77        & --             & 2.17          & --           & --         \\
                     & Ours $(10,10)$          & $2.78$           & $2.31^{\pm 0.01}$        & 2.89           & $2.50^{\pm 0.01}$             & ${2.56}$            & $2.21^{\pm 0.01}$         \\
                     & Ours $(k^*, w^*)$  & 2.68           & $2.45^{\pm 0.02}$         & 2.91              & $2.77^{\pm 0.02}$             & 2.46            & $2.32^{\pm 0.01}$          \\\bottomrule
\end{tabular}
\vspace{0.4em}
\caption{}
\label{table:results}
\vspace{-0.5em}
\end{table}

For each dataset and model, we report the strategy that led to the largest wall-time speedup, which we denote $(k^*, w^*)$. As a reference, we also reported $(k, w) = (10, 10)$, to compare how a square input block, representative of a default non-optimized choice in our sweep fared compared to $(k^*, w^*)$. Table~\ref{table:results} shows the mean and standard deviation across the three runs, for all models and datasets. 

\begin{figure}[!b]
    \centering
    \begin{subfigure}{0.32\textwidth}
        \includegraphics[width=\linewidth]{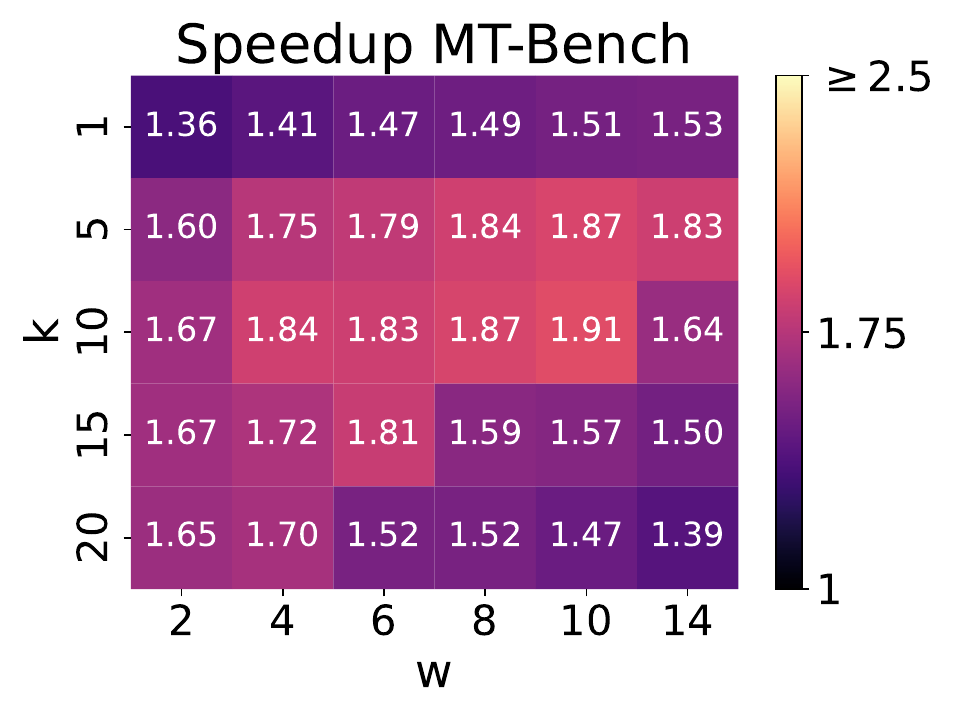}
    \end{subfigure}
    \begin{subfigure}{0.32\textwidth}
        \includegraphics[width=\linewidth]{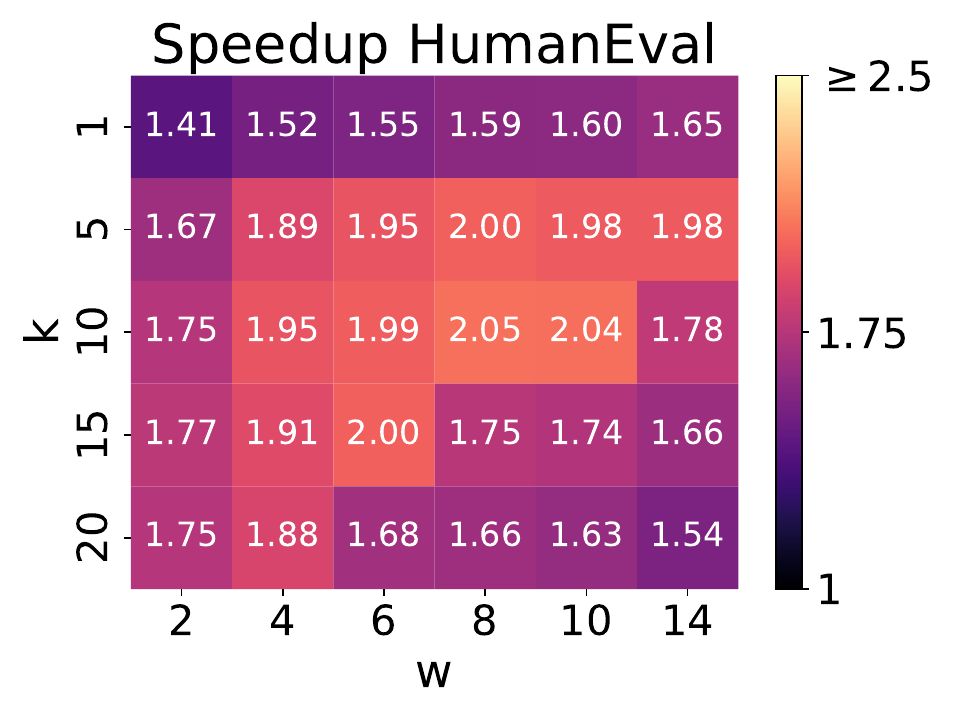}
    \end{subfigure}
    \begin{subfigure}{0.32\textwidth}
        \includegraphics[width=\linewidth]{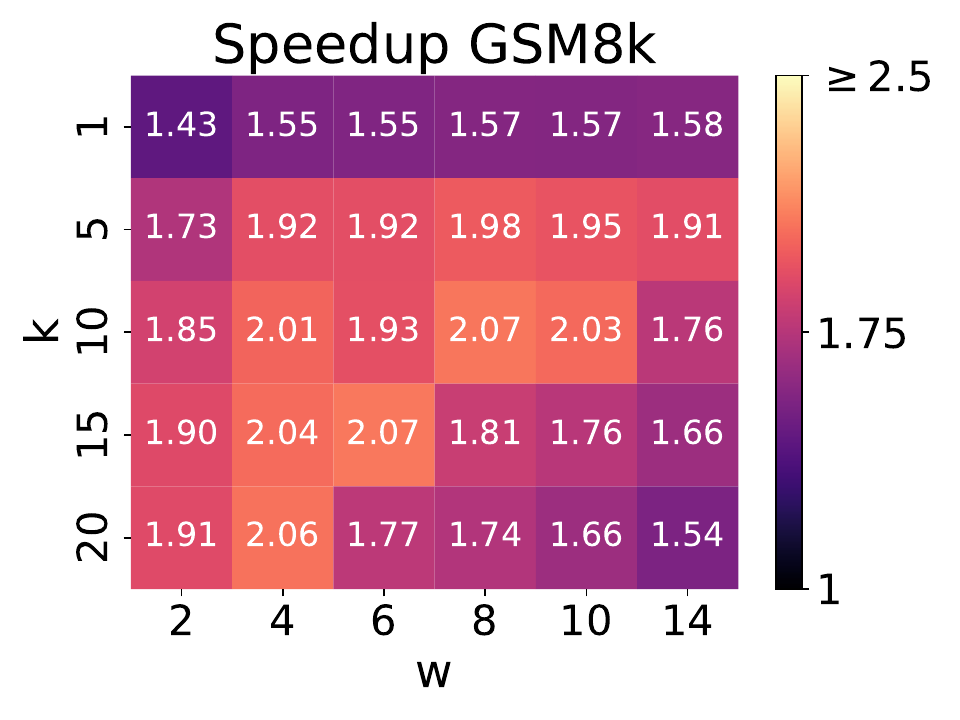}
    \end{subfigure}
    \caption{Average wall-time speedup across datasets for Mistral7B instruct for varied $(k, w)$.}
    \label{fig:speedup_grid_m7b}
\end{figure}

For Mistral7B, the average wall-time speedups for the complete grid of strategies is depicted in Figure ~\ref{fig:speedup_grid_m7b}. The grids all show a clear tradeoff between tokens-per-call (by increasing either $k$ and/or $w$) and compute-bound slowdowns, with the pattern of Figure~\ref{fig:speedup_grid_m7b} shared across the three different tasks suggesting a consistent relationship between $(k ,w)$ and speed-up. For reference, the corresponding tokens per call are reported in Appendix \ref{app:m7btpca}. The equivalent plots for both Phi3B and Vicuna13B can be found in Appendix~\ref{app:additionalres}. For Phi3B, we notably observed the model never reached an OTP ratio that was large enough to incur slowdowns which would outweigh increases in tokens per call. For this reason the optimal values were trivially those of maximum value, \ie $(k^*,w^*) = (25, 14)$ ;  the true optimal speed-up using our batched approach would hence occur at a larger $(k, w)$.

Overall, our methods consistently achieve more than $2$x speedup across models and tasks (except for the 7B model on MT bench, which attained a 1.91 times speedup). While the optimal $(k^*, w^*)$ varied between models and datasets, it can be seen that the representative default $(10, 10)$ achieved good performance on all of the settings. 

\begin{figure}[t]
    \centering
    \includegraphics[width=0.9\textwidth]{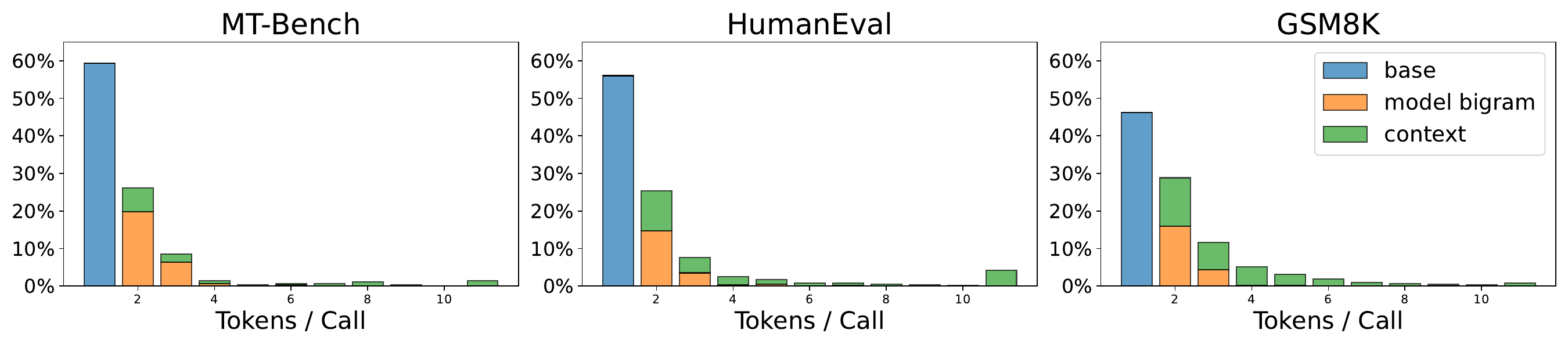}\\
    \includegraphics[width=0.9\textwidth]{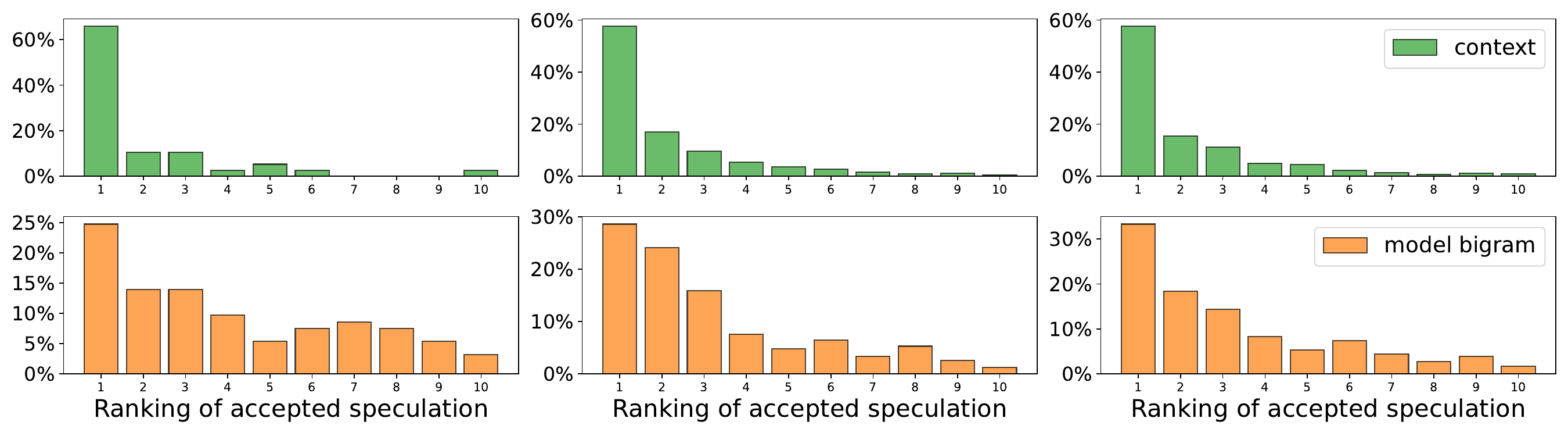}\\
    \includegraphics[width=0.9\textwidth]{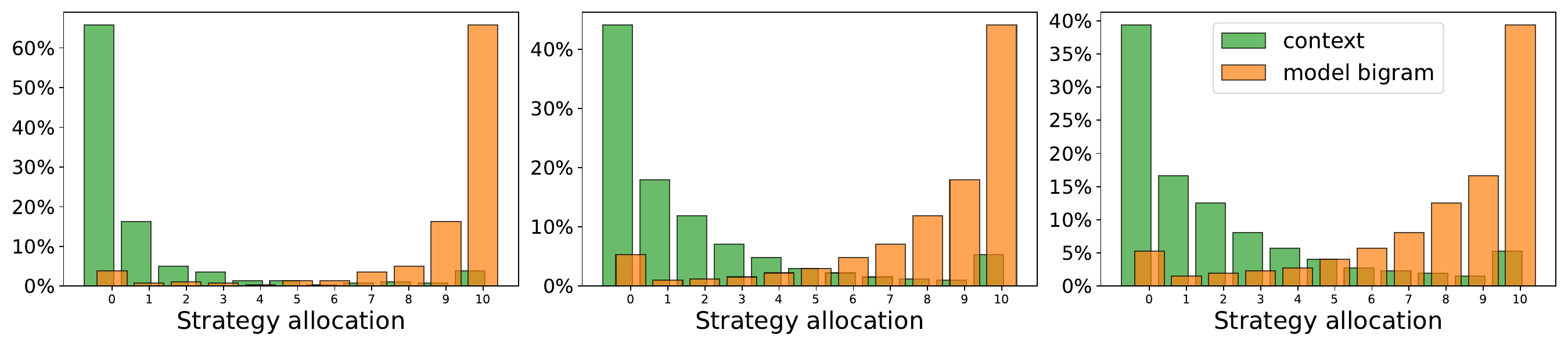}
    \caption{Ablations: \textbf{Top}: distribution of acceptance length for mixed strategies. \textbf{Middle}: distribution of ranking of accepted speculations amongst the top-$k$. \textbf{Bottom:} allocation distribution of strategies \ie number of speculations for each strategy.}
    \label{fig:ablations}
\end{figure}

\subsection{Ablation on strategies}\label{subsec:abl}

In order to understand the role that both the model- and context-derived N-grams play in the observed speed-ups, we ablate the Mistral7B experiment for $(k,w)=(10,10)$ across the three data sets. We explore i) the number of speculated tokens accepted by both the model and context derived N-grams ii) the rank of accepted speculations amongst the top-$10$ speculations iii) the amount of drafts \ie rows in the batch, that each of the strategies used. The results are depicted in Figure~\ref{fig:ablations}.

Our ablations shine light on the strengths and weaknesses of both the model bigram and context derived draft strategies. The model-bigram is robust across all tasks, with an additional 1-2 future tokens frequently found within its top-$10$ predictions, with the ranking distribution being notably heavy-tailed (relative to that of the context-derived N-gram). The model's bigram weakness lies in its ineffectiveness for larger values of $w$, which is expected since it only considers the last token of the context rather than encompassing all prior context, making it insufficient for longer speculations. 

On the other hand, we see that the context derived N-gram compliments the bigram's weakness as it can successfully speculate further into the future, with speculations of length $w=10$ accepted on all tasks. However, its performance is notably less robust across tasks. For example, GSM8K exhibits a wider distribution of accepted lengths due to the varied sizes of calculations in the math word-problems, while HumanEval more frequently accepts $w=10$ length speculations due to the coding nature of the task. Furthermore, it exhibits more pronounced diminishing returns from batching (compared to the bigram), which is a weakness given that it is often allocated the entire batch for speculations (see the bottom row of Figure \ref{fig:ablations}). This suggests that further research into enhancing strategy allocation could indeed yield further additional gains.

\section{Conclusions}

We introduced a set of learning-free strategies for generating batches of speculative drafts, extracted from both the base model and context. Our approach is conceptually simple and is fully compatible with other optimization techniques (e.g. quantization, early exiting, flash attention, etc.). Experimentally, we observed that our proposed strategies led to significant speedups in auto-regressive inference, while requiring minimal implementation overhead and being easily integrable. Our analysis demonstrates that simple strategy combinations can substantially enhance performance across a range of different tasks and model sizes, with our ablations shining light on the strengths and weaknesses of the proposed strategies.

\newpage

\bibliographystyle{plainnat}
\bibliography{references}

\newpage

\appendix

\section{Additional results}\label{app:additionalres}
\subsection{Mistral 7B}\label{app:m7btpca}

\begin{figure}[!ht]
    \centering
    \begin{subfigure}{0.32\textwidth}
        \includegraphics[width=\linewidth]{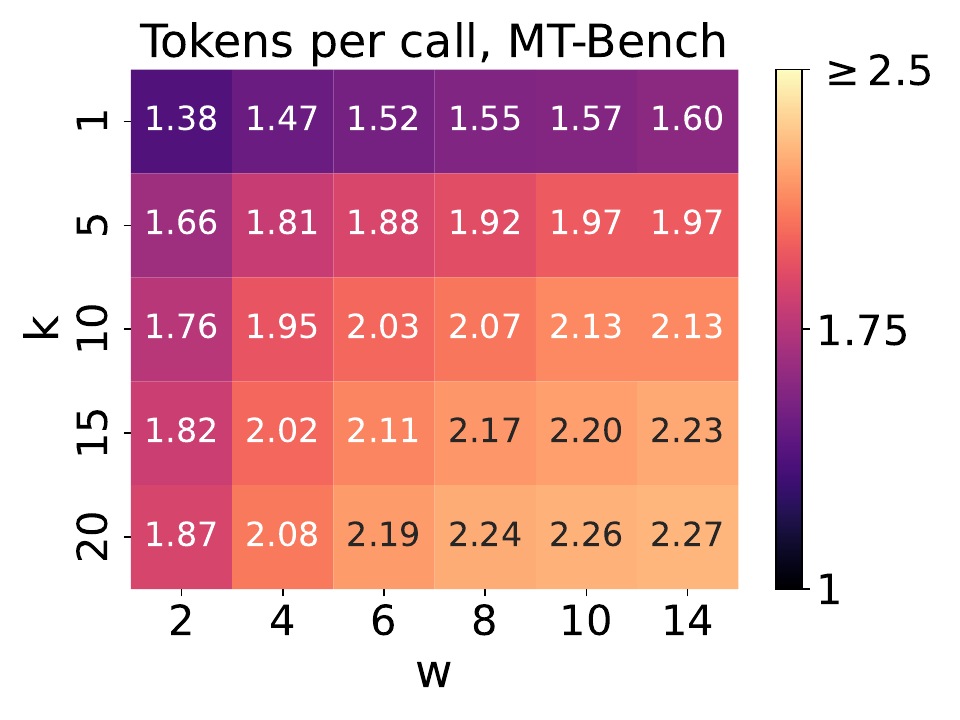}
    \end{subfigure}
    \begin{subfigure}{0.32\textwidth}
        \includegraphics[width=\linewidth]{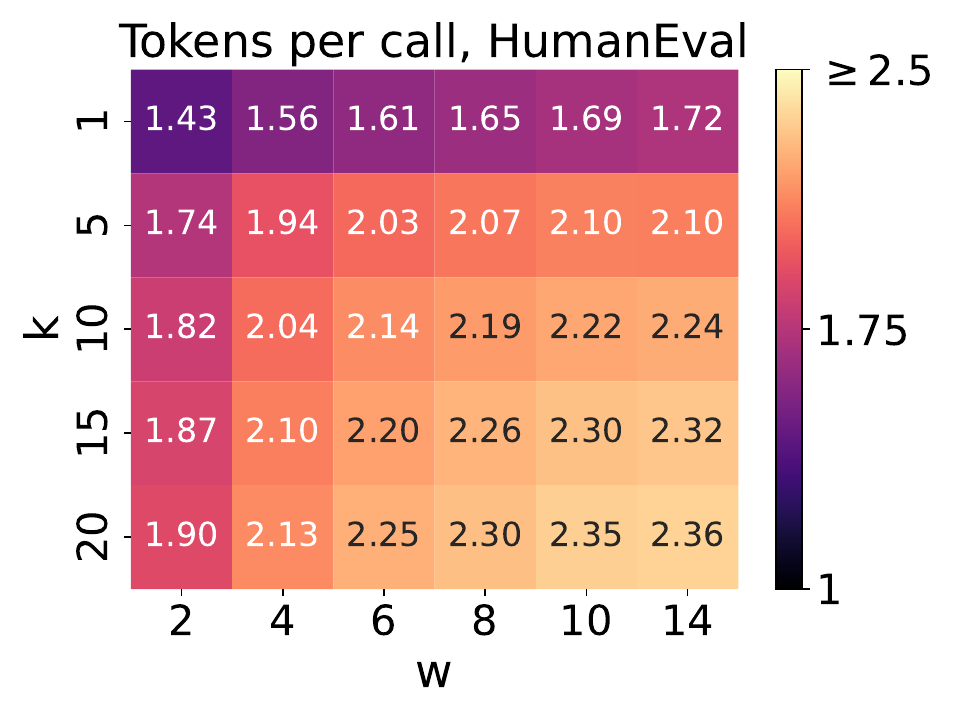}
    \end{subfigure}
    \begin{subfigure}{0.32\textwidth}
        \includegraphics[width=\linewidth]{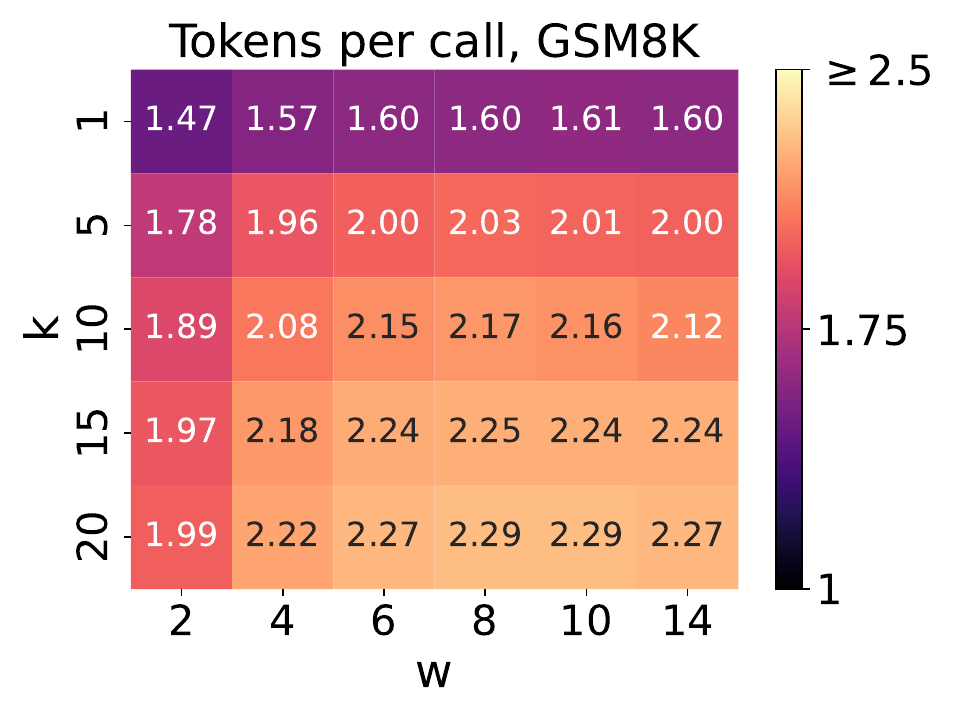}
    \end{subfigure}
    \caption{Tokens per call across datasets for Mistral7B instruct for varied $(k, w)$.}
    \label{fig:tpc_grid_m7b}
\end{figure}
\subsection{Phi3B}
We remark that Phi3B never was compute-bound, so maximum speed-up from mixing strategies was not attained.

\begin{figure}[!ht]
    \centering
    \begin{subfigure}{0.32\textwidth}
        \includegraphics[width=\linewidth]{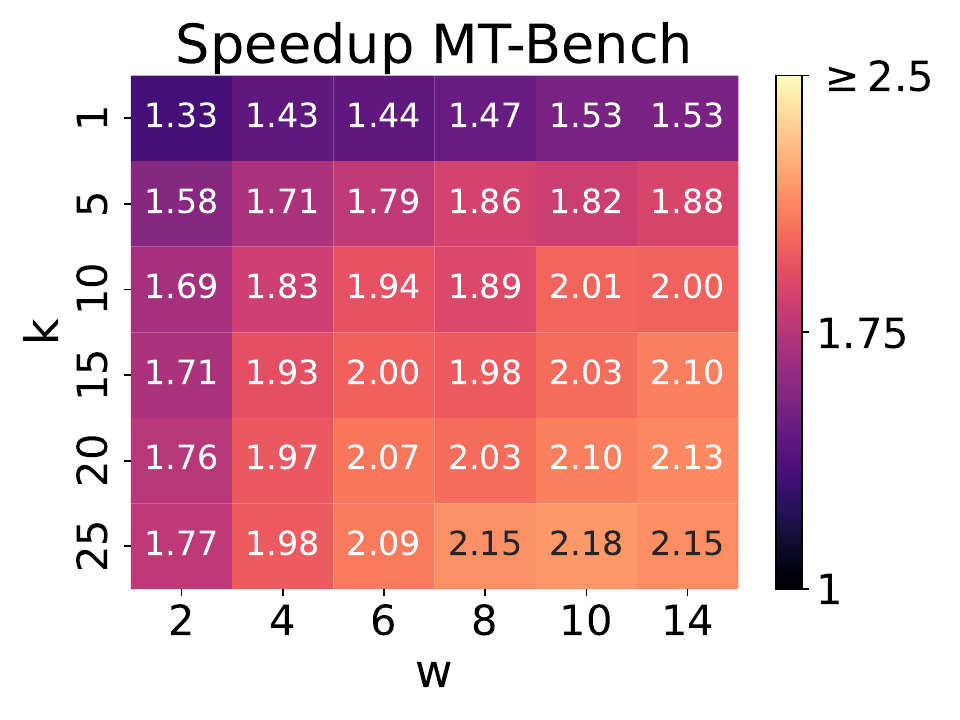}
    \end{subfigure}
    \begin{subfigure}{0.32\textwidth}
        \includegraphics[width=\linewidth]{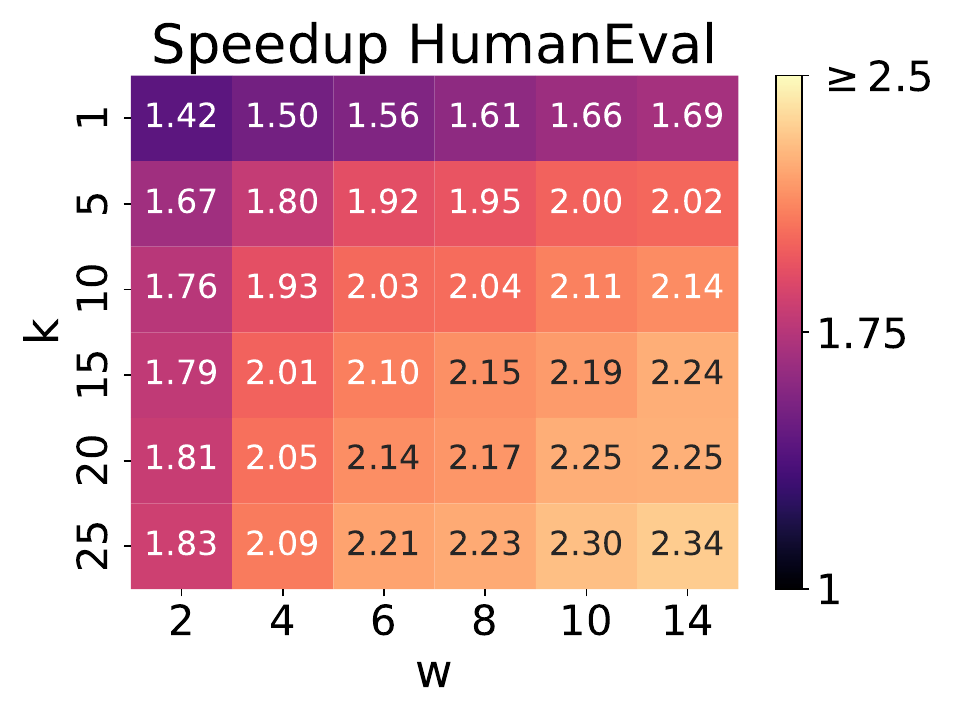}
    \end{subfigure}
    \begin{subfigure}{0.32\textwidth}
        \includegraphics[width=\linewidth]{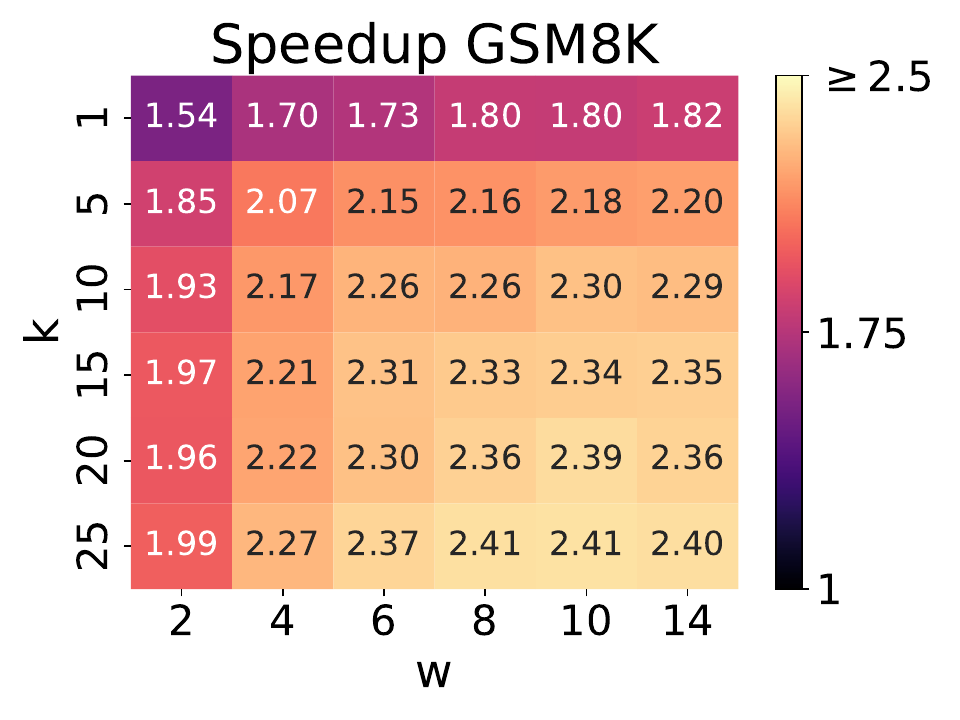}
    \end{subfigure}
    \caption{Average wall-time speedup across datasets for Phi3B-instruct for varied $(k, w)$.}
    \label{fig:speedup_grid_phi}
\end{figure}

\begin{figure}[!ht]
    \centering
    \begin{subfigure}{0.32\textwidth}
        \includegraphics[width=\linewidth]{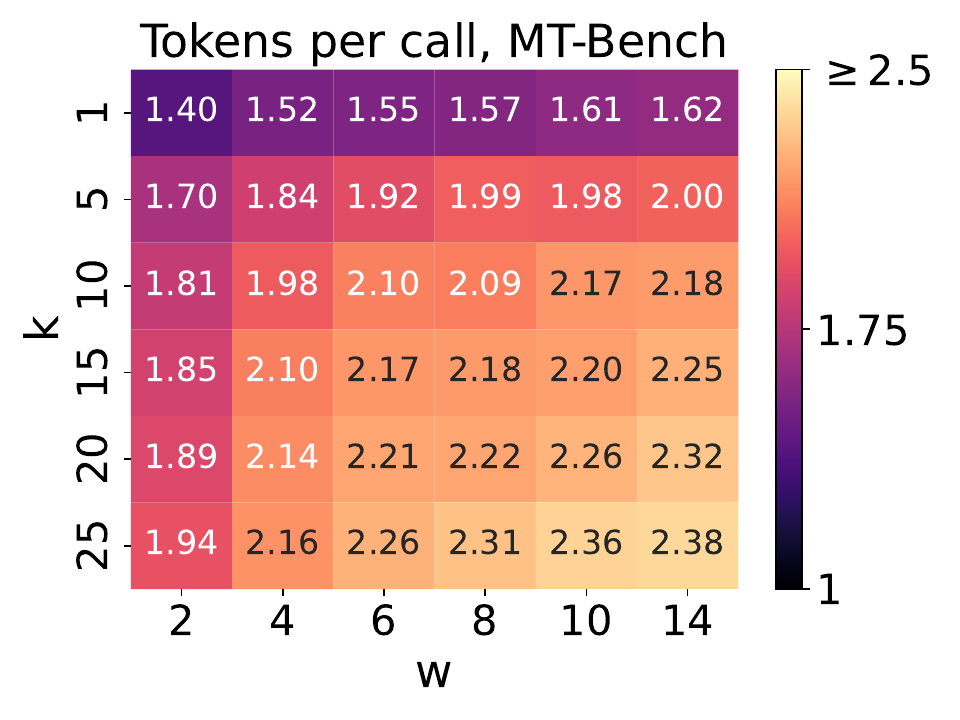}
    \end{subfigure}
    \begin{subfigure}{0.32\textwidth}
        \includegraphics[width=\linewidth]{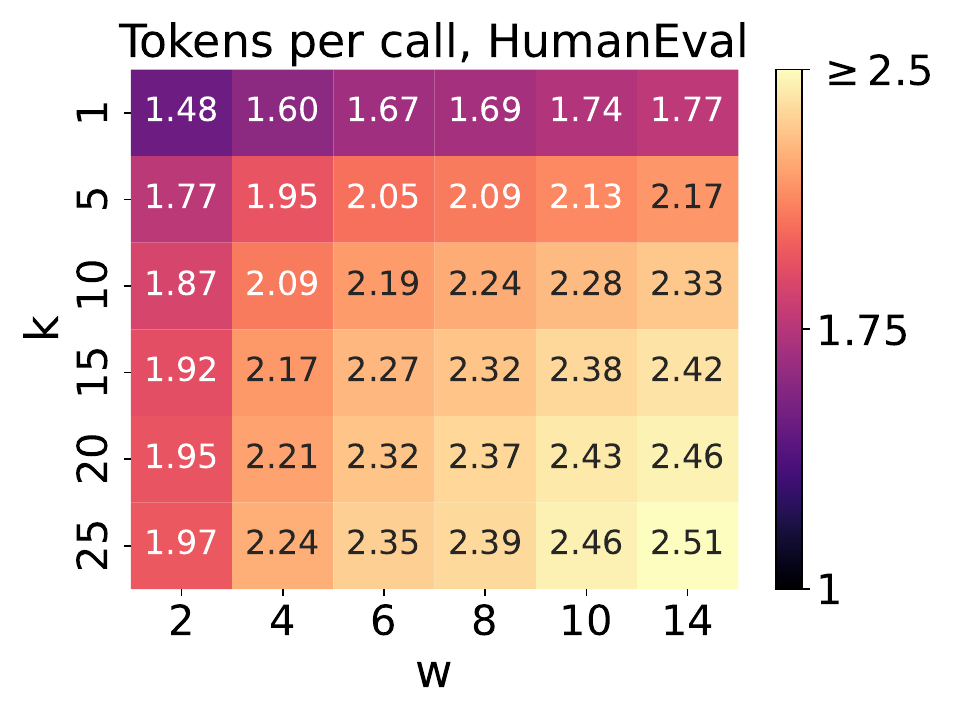}
    \end{subfigure}
    \begin{subfigure}{0.32\textwidth}
        \includegraphics[width=\linewidth]{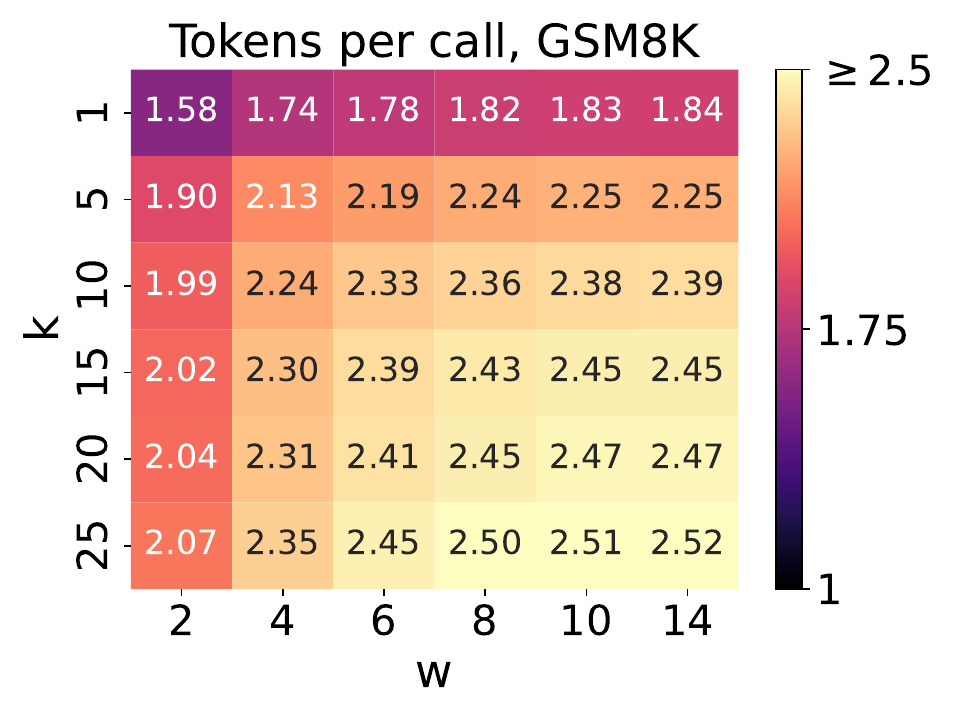}
    \end{subfigure}
    \caption{Tokens per call across datasets for Phi3B-instruct for varied $(k, w)$.}
    \label{fig:tpc_grid_phi}
\end{figure}

\subsection{Vicuna 13B}

\begin{figure}[!ht]
    \centering
    \begin{subfigure}{0.32\textwidth}
        \includegraphics[width=\linewidth]{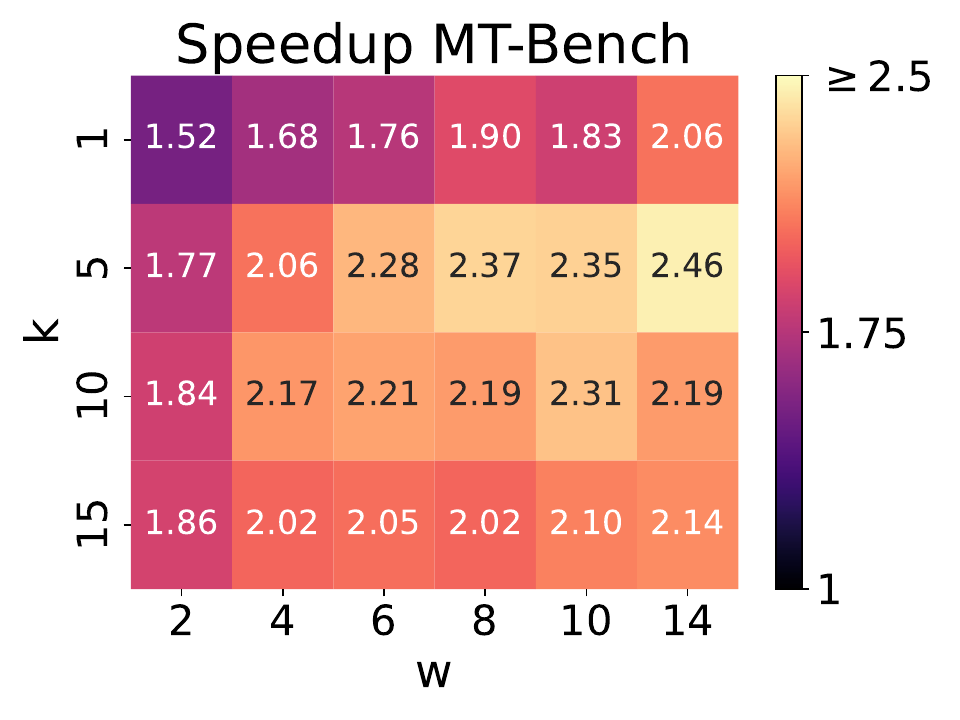}
    \end{subfigure}
    \begin{subfigure}{0.32\textwidth}
        \includegraphics[width=\linewidth]{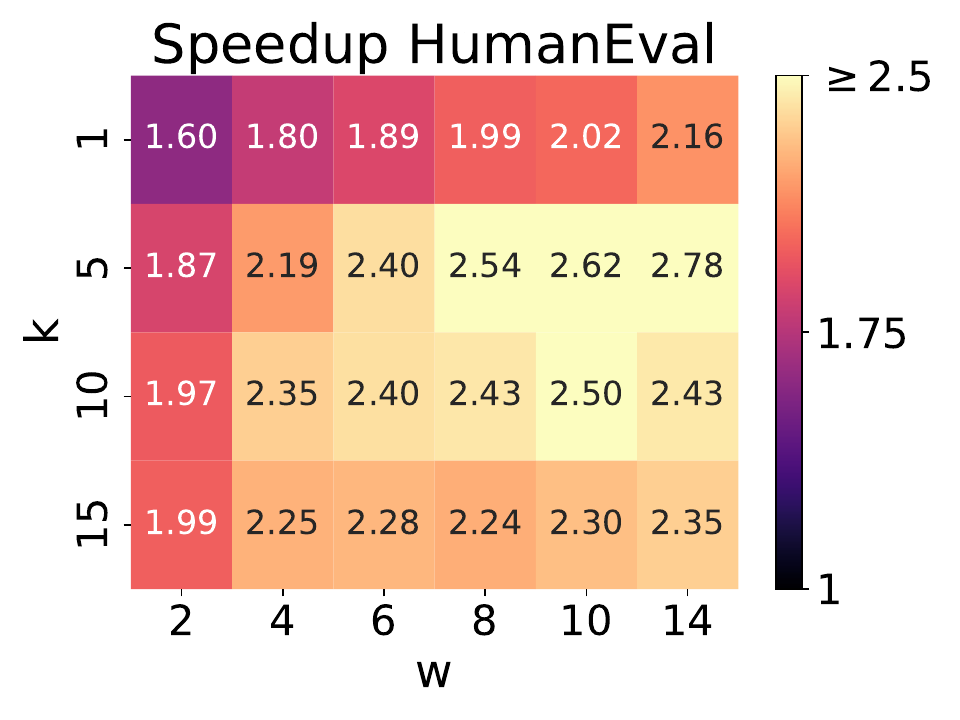}
    \end{subfigure}
    \begin{subfigure}{0.32\textwidth}
        \includegraphics[width=\linewidth]{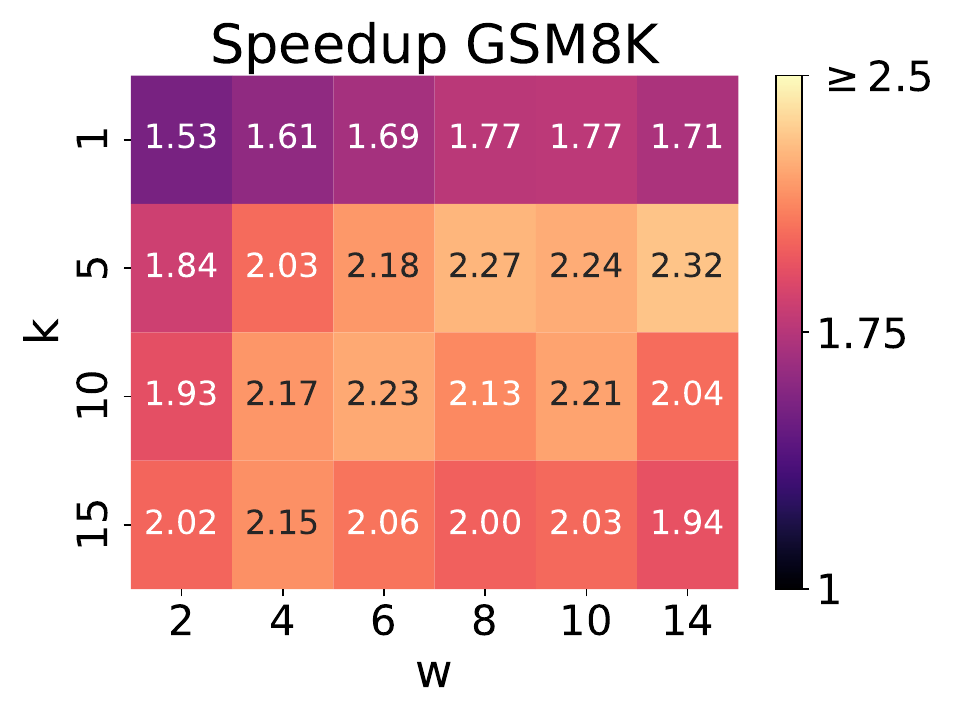}
    \end{subfigure}
    \caption{Average wall-time speedup across datasets for Vicuna13B for varied $(k, w)$.}
    \label{fig:speedup_grid_vicuna}
\end{figure}

\begin{figure}[!ht]
    \centering
    \begin{subfigure}{0.32\textwidth}
        \includegraphics[width=\linewidth]{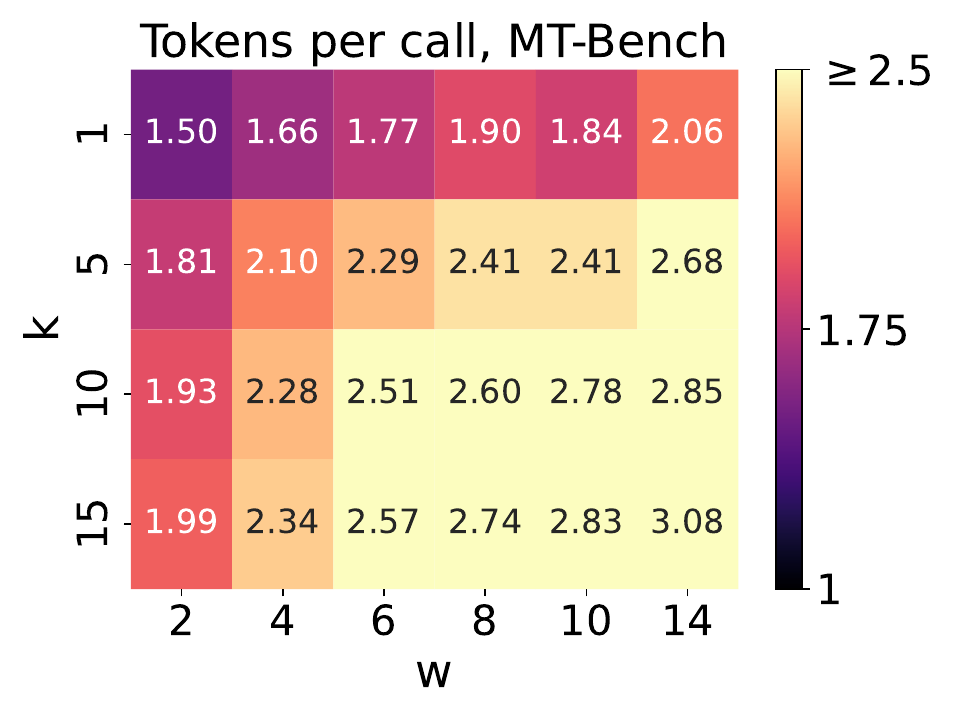}
    \end{subfigure}
    \begin{subfigure}{0.32\textwidth}
        \includegraphics[width=\linewidth]{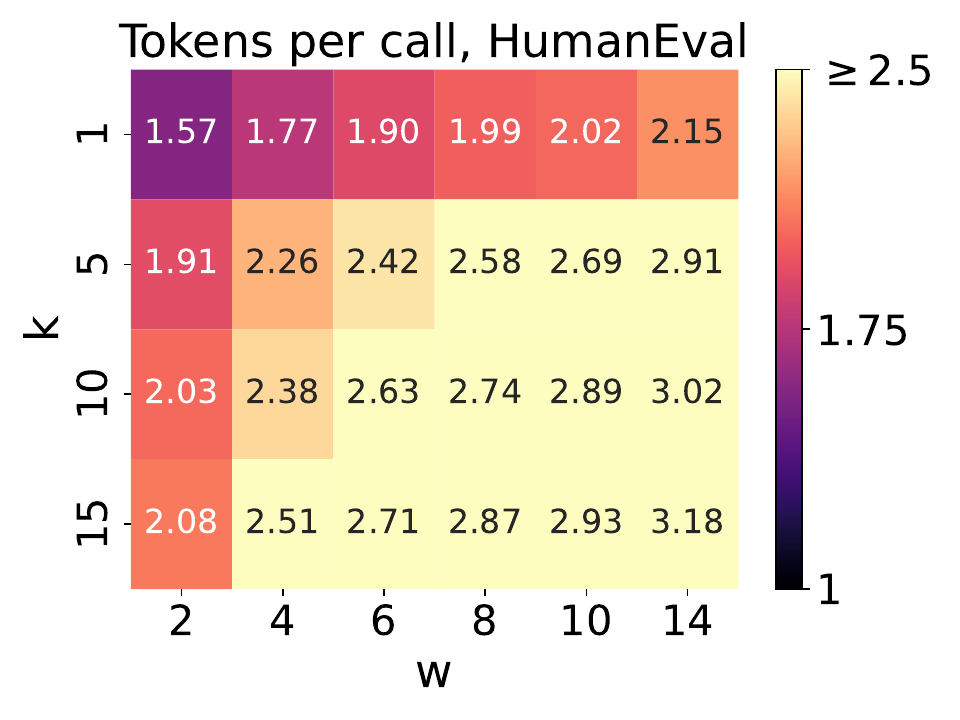}
    \end{subfigure}
    \begin{subfigure}{0.32\textwidth}
        \includegraphics[width=\linewidth]{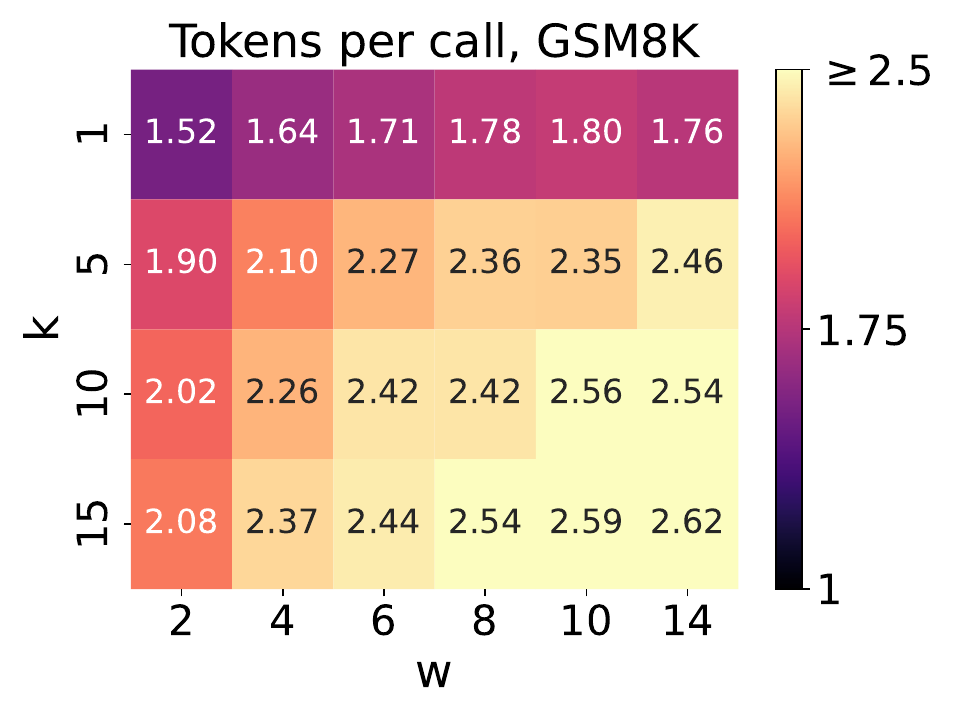}
    \end{subfigure}
    \caption{Tokens per call across datasets for Vicuna13B for varied $(k, w)$.}
    \label{fig:tpc_grid_vicuna}
\end{figure}

\newpage

\section{Learning-Free N-Grams}\label{app:Ngrams}
\subsection{Model Based N-Grams}\label{app:modelNgrams}

\begin{lstlisting}
def unigram(model):
    '''
    Obtain a Unigram topk from a transformer's weights
    '''
    Wenc = model.get_input_embeddings().weight.detach()
    covV = Wenc.T @ Wenc / Wenc.shape[0]
    Wdec = model.lm_head.weight.detach()

    # look at the distance from the mean embedding in the decoder space.
    mu = Wdec.mean(dim=0, keepdim=True)
    dists = mu @ covV @ Wdec.T
    dists = dists.squeeze()
    ranks = torch.topk(-dists, k=dists.size(0)).indices

def bigram(model):
    '''
    Pseudo code: bigram topk from a transformer:
    '''
    V = model.config.vocab_size

    lookup = torch.empty(V, V)

    # this can be done in batches
    for x in range(len(V)):
        lookup[x] = model(x).logits
    return lookup

\end{lstlisting}

\subsection{Context Based N-Grams}\label{app:contextNgrams}
\begin{lstlisting}
@torch.inference_mode
def context_ngram_matcher(context, query, Ndraft=1, Npad=1):
    '''
    matches query of any length Q to grams of size Q + Npad

    query : tensor (query to match with)
    Ndraft : int (number of drafts)
    Npad : int (number of tokens to speculate with)
    '''
    # obtain length of query
    Q = query.size(-1)
    # use unfold to obtain all N grams
    grams = context.flatten().unfold(0, Q + Npad, 1)

    # extract mask of matching ngrams
    mask = torch.all(grams[:, :Q] == query, dim=-1)
    if torch.any(mask):
        matching_grams = grams[mask]
        # obtain counts of all ngrams
        matches, counts = torch.unique(matching_grams, dim=0, return_counts=True)
        Nfound = counts.size(0)
        Ntake = min(Ndraft, Nfound)
        # take up to top Ndraft occuring Ngrams
        most_freq_ids = counts.topk(Ntake).indices
        return matches[most_freq_ids]
    else:
        return None
\end{lstlisting}

\section{Models}\label{app:modelurls}

\begin{itemize}
    \item Phi3B : \url{https://huggingface.co/microsoft/Phi-3-mini-4k-instruct} (MIT License)
    \item Mistral 7B : \url{https://huggingface.co/mistralai/Mistral-7B-Instruct-v0.2} (Apache 2.0 License)
    \item Vicuna 13B : \url{https://huggingface.co/lmsys/vicuna-13b-v1.3} (Non-commercial license)
\end{itemize}

\section{Key-Value Cache}

We use a static key-value cache based upon the implementation from \citet{cai2024medusa, li2024eagle}. However, we add minimal modifications to i) allow for batching ii) over-write all rows to be that of the maximum length accepted speculation iii) initialize from a $k=1$ cache (since the context is repeated), via a broadcasting.

\newpage

\end{document}